\documentclass[runningheads]{llncs}
\usepackage[T1]{fontenc}
\usepackage{graphicx}
\begin{document}
%
\title{Influence of Backdoor Paths on Causal Link Prediction}
%
%
\author{Utkarshani Jaimini\inst{1}\orcidID{0000-0002-1168-0684} \and
Cory Henson\inst{2}\orcidID{0000-0003-3875-3705} \and
Amit Sheth\inst{1}\orcidID{0000-0002-0021-5293}}

\authorrunning{F. Author et al.}
%
\institute{Artificial Intelligence Institute, University of South Carolina, Columbia, SC, USA \email{ujaimini@email.sc.edu, amit@sc.edu} \and
Bosch Center for Artificial Intelligence, Pittsburgh, PA, USA 
\email{cory.henson@us.bosch.com}
}
\maketitle              
\begin{abstract}

The current method for predicting causal links in knowledge graphs uses weighted causal relations. For a given link between cause-effect entities, the presence of a confounder affects the causal link prediction, which can lead to spurious and inaccurate results. We aim to block these confounders using backdoor path adjustment. Backdoor paths are non-causal association flows that connect the \textit{cause-entity} to the \textit{effect-entity} through other variables. Removing these paths ensures a more accurate prediction of causal links. This paper proposes CausalLPBack, a novel approach to causal link prediction that eliminates backdoor paths and uses knowledge graph link prediction methods. It extends the representation of causality in a neuro-symbolic framework, enabling the adoption and use of traditional causal AI concepts and methods. We demonstrate our approach using a causal reasoning benchmark dataset of simulated videos. The evaluation involves a unique dataset splitting method called the Markov-based split that's relevant for causal link prediction. The evaluation of the proposed approach demonstrates atleast 30\% in MRR and 16\% in Hits@K inflated performance for causal link prediction that is due to the bias introduced by backdoor paths for both baseline and weighted causal relations. 

\end{abstract}

\section{Introduction}

\begin{figure}
    \centering
\includegraphics[width=0.8\textwidth, height=0.4\textwidth]{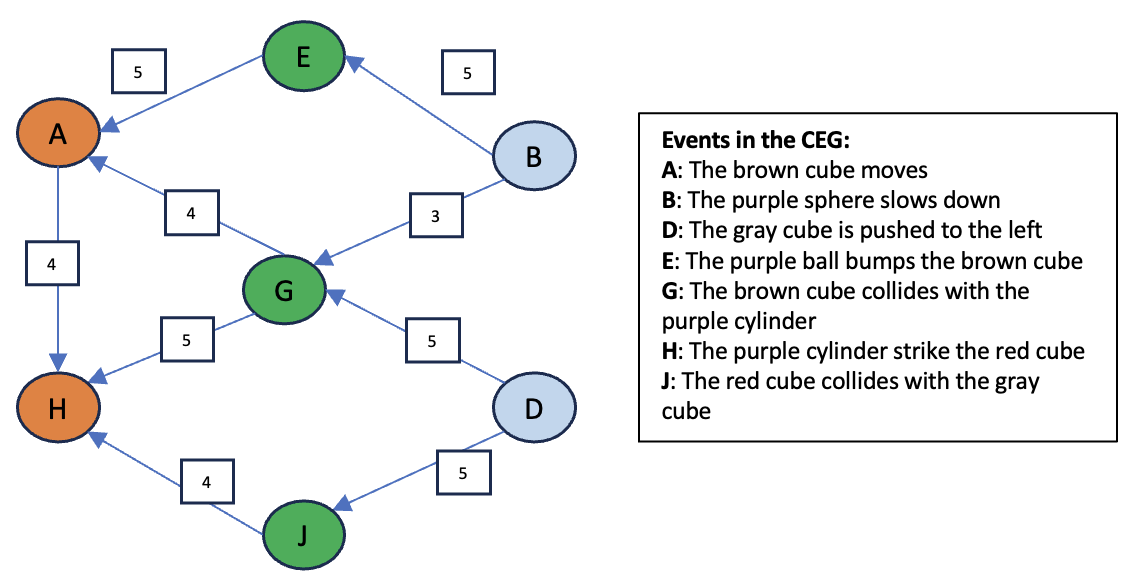}
    \caption{A causal network where the nodes represent the event and the edges between the nodes represent the causal relations. To estimate the true causal influence of node A on H (marked in orange) we need to block the influence of the common cause node, G which is affecting both A and H. Hence the edges on the path from A to H going via G is removed from the network      
}
    \label{fig:BDExample}
\end{figure}

Knowledge graph link prediction is the process of discovering new knowledge using the existing knowledge represented in the knowledge graph (KG). Causal link prediction is the process of discovering new causal links using the existing knowledge represented in the knowledge graph (KG).  
The use of knowledge graph link prediction for causal link prediction is a first step towards the use KGs (and neuro-symbolic AI methods) for causal AI tasks.
However, the existing efforts on causal link prediction using KG link prediction do not consider the presence of backdoor paths for a given causal link \cite{CausalDisco} even though the backdoor path demonstrates a common cause of the entities present in the causal link. The presence of a backdoor path in the causal link prediction can lead to inflated evaluation results due to the bias (spurious correlations) introduced by the backdoor paths \cite{Whypearl2018book,pearl2009causality}. Figure \ref{fig:BDExample} illustrates an example of backdoor path in a causal network. This paper proposes a novel approach, CausalLPBack, for finding new causal relations using KG link prediction while also removing the backdoor paths to mitigate the backdoor induced bias. It further extends the representation of causality in the KG, enabling the adoption and use of traditional causal AI concepts and methods (see causal neuro-symbolic AI \cite{jaimini2024causalNSAI}). The paper utilizes the formulation of the causal link prediction task using KG link prediction \cite{CausalDisco}.  
Causal networks (CN) are often incomplete, due to missing observation data. CausalLPBack encodes the incomplete CN as a KG, framing the task of finding missing causal relations as a KG completion problem, i.e., identifying missing links in the KG. The link prediction model in this paper is trained on known causal links within the CN to infer new causal links. The initial CN structure can also be provided by domain experts or learned from data. 
CausalLPBack is a versatile framework that’s applicable to real-world applications, from causal  explanation in autonomous driving (AD), to root cause analysis in smart manufacturing, to identifying the cause of symptoms in multifactorial diseases.  It is adaptable to any domain and KG with causal associations. 

The CausalLPBack architecture is comprised of five primary phases: (1) translating known causal relations into a causal network, (2) removing backdoor path between the testing and training set, (3) transforming the causal network into a knowledge graph, (4) learning knowledge graph embedding models for the causal relations, and (5) predicting new causal links in the knowledge graph. The KG captures causality in a causal link with the form $<$\textit{cause-entity, \textit{\textbf{\textit{causes}}}, effect-entity, w}$>$, where the \textit{\textbf{causes}} relation links the cause and effect entities. The \textit{causes} relation is associated with a \textit{causal weight},  \textit{w}, representing the causal influence of the \textit{cause-entity} on the \textit{effect-entity}. A KG embedding (KGE) that incorporates weighted relations is used to learn new causal relations \cite{focusE}. The task of causal link prediction using KG link prediction is divided into two sub tasks: causal prediction and causal explanation.  When implemented with KG link prediction, causal explanation is mapped to the task of finding the type of the head (i.e. \textit{cause-entity}) of a causal link, and causal prediction is mapped to the task of finding the type of the tail (i.e. \textit{effect-entity}) of a causal link.  For the evaluation, we utilize the Markov-based data split which follows the Markovian property of the causal network to mitigate the issues of data leakage and model bias \cite{CausalDisco}. The main contributions of the paper are
\begin{enumerate}
    \item A novel approach for finding new causal links using knowledge graph link prediction without the backdoor paths.
    \item Demonstration of the approach on a causal benchmark dataset of simulated videos. 
    \item Evaluation of the approach using multiple KG embedding models which reveals improved performance with the use of weighted causal relation.
    \item Illustration of the bias in the causal link prediction due to the presence of backdoor paths.
\end{enumerate}
The rest of the paper proceeds as follows: Section 2 discusses related work. Section 3 defines the problem formulation followed by methodology in Section 4. Section 5 discusses the experiment. Section 6 outlines the results and discussion. Section 7 provides a conclusion with future direction.

\section{Related work}

\textbf{Causal knowledge graphs}: CauseNet is a knowledge graph that captures causal concepts and relationships derived from both semi-structured and unstructured web sources \cite{heindorf2020causenet}. In CauseNet, causal relationships are represented as <subject may-Cause object> links. To identify these relationships, linguistic patterns are employed, focusing on words like "cause(d)", "result(ed)", "lead(s)/led", and "associated with" to define causal statements. These statements are assumed to contain causal concepts, such as cause and effect, which are treated as nouns, akin to the approaches used in ConceptNet and WordNet. While the linguistic patterns effectively extract explicit causal relationships, they do not capture implicit causal mentions or relationships that span multiple sentences. Additionally, since the data is sourced from the web, it can reflect societal biases. Examples of biased causal relationships found in the extracted data include claims like "Autism is caused by vaccination" and "HIV is caused by homosexuality" \cite{heindorf2020causenet}.

\textbf{Causal link prediction}: In the domain of causal link prediction, existing techniques typically focus on predicting general links within knowledge graphs and lack specific tailoring for identifying causal links. However, recent work has aimed to generate event-related causal knowledge graphs from sources like Wikipedia and Wikidata, incorporating causal predicates like hasCause and hasEffect \cite{hassanzadeh2022causalbuilding}. These graphs represent events as nodes and cause-effect relationships as edges, with the objective of predicting future events by analyzing the underlying causes and effects of similar past events. Evaluation of causal link prediction tasks often utilizes established techniques for knowledge graph link prediction. This recent work lays the foundation for causal link prediction \cite{CausalDisco}.  However, CausalLPBack further infuses the KG based causal link prediction methods with causal concepts such as backdoor path leading to a better framework and methodology to perform causal AI tasks.


\section{Problem formulation}


\textbf{Causal knowledge graph:}
Causal knowledge is structured within a causal knowledge graph, denoted as $CausalKG$, which is comprised of causal entities, causal relations, and causal weights. The structure of $CausalKG$ ($N$, $R$, $E$, $E_c$, $W_c$) is detailed as follows:

\begin{itemize}
    \item $N$: a set of nodes that represent entities.
     \item $R$: a set of labels that denote different types of relations.
     \item $E$ $\subseteq$ $N \times R \times N$:  a set of edges that form links between pairs of entities, with each link represented as a triple <$h$, $r$, $t$>, where $h$ is the head entity, $r$ is the relation, $t$ is the tail entity.
     \item $N_c \subseteq N$: a subset of nodes that are identified as causal entities.
     \item $R_c \subseteq R$: a subset of labels that are identified as causal relations.
     \item $W_c \subseteq R$: a set of real numbers that assign causal weights.
     \item $E_c$ $\subseteq$ $N_c \times R_c \times N_c \times W_c$:  a set of edges that connect pairs of causal entities. Each of these causal links is represented as a quadruple <$h_c$, $r_c$, $t_c$, $w_c$>, where $h_c$ is the head causal entity, $r_c$ is the causal relation, $t_c$ is the tail causal entity, and $w_c$ is the weight of the causal influence.
 \end{itemize}
This graph encapsulates the causal structure and dynamics between entities, offering a clear depiction of causal interactions within the system.

\textbf{Causal entity:}
A causal entity {$n_c \in N_c$} is defined as any entity that serves as either the head or the tail in a causal link. There are two distinct categories of causal entities: the \textit{cause-entity} ($n_{cause}$)  and the \textit{effect-entity} ($n_{effect}$). In a causal relation, the \textit{cause-entity} is responsible for causing the \textit{effect-entity}.

\textbf{Causal relation:}
A causal relation {$r_c \in R_c$} denotes a relation that represents a causal linkage between entities. There are four specific types of causal relations:
\begin{itemize} 
\item  $causes$ ($r_{causes} \in R_c$)  is a causal relation that links a cause-entity to an effect-entity.
\item  $causedBy$ ($r_{causedBy} \in R_c$)  is a causal relation that connects an effect-entity back to a cause-entity; the reverse of {$causes$}.
\item $causesType$ ($r_{causesType} \in R_c$) is a causal relation that associates a cause-entity with the type of the effect-entity.
\item $causedByType$ ($r_{causedByType} \in R_c$)  is a causal relation that links an effect-entity to the type of the cause-entity. 
\end{itemize}
These relations categorize the directional and typological aspects of causality within the $CausalKG$.

\textbf{Causal weight:}
A real number causal  weight $w \in W_c \subseteq R$  associated with a causal link. It quantifies the responsibility or contribution of the cause-entity in causing the effect-entity.

\textbf{Causal link:}
A causal link $e_c \in E_c$ is an edge within the $CausalKG$ that connects two causal entities via a specified causal relation, accompanied by a particular causal weight. This causal link is defined as a quadruple <$h_c$, $r_c$, $t_c$, $w_c$>, where  {$h_c$} represents the head causal entity, $r_c$ denotes the causal relation, $t_c$ is the tail causal entity, and $w_c$ is the weight that quantifies the causal influence.

\textbf{Causal Link Prediction:}
Causal link prediction is the task of identifying new causal links within a $CausalKG$. Utilizing KG link prediction techniques, this task can be accomplished for a given CausalKG, \(G\). There are two primary forms of causal link prediction: causal prediction and causal explanation.
\begin{enumerate}
    \item Causal Prediction: the task of identifying the type ($t$) of the effect-entity associated with a given cause-entity ($n_{cause} \in N_c$) and the \(causesType\) relation ($r_{causesType} \in R_c$). The objective is to find a match such that $<$$n_{cause}$, $r_{causesType}$, $t$, $w_c$$>$ is a valid link in \(G\).
    \item Causal Explanation: the task of identifying the  type ($t$) of the cause-entity associated with a given effect-entity ($n_{effect} \in N_c$) and the \(causedByType\) relation  ($r_{causedByType} \in R_c$). The objective is to find a match such that $<$$n_{effect}$, $r_{causedByType}$, $t$, $w_c$$>$ is a valid link in \(G\).
\end{enumerate}

\textbf{Backdoor path:} 
A backdoor path, $BD$, is a non-causal path  between a \textit{cause-entity} ($n_{cause}$) and its \textit{effect-entity} ($n_{effect}$). A backdoor path contains an incoming edge at both the \textit{cause-entity} and the \textit{effect-entity}.  

\begin{figure*}
 \centering
\includegraphics[width=\textwidth, height=0.42\textwidth]{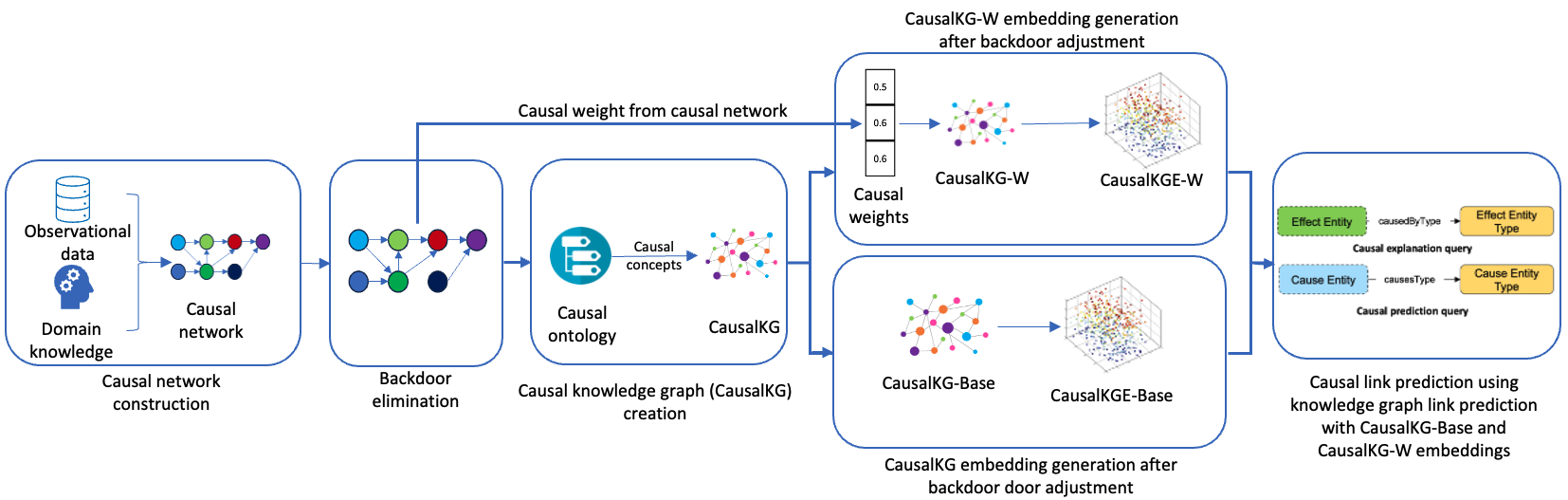}
 \caption{CausalLPBack has five primary phases: 1) encoding the causal associations from data as a causal network, 2) removing the backdoor paths spanning across the nodes in the training and testing set (3) translating the causal network into a CausalKG, compatible with the causal ontology,   (4) learning KGE for the CausalKG-W and CausalKG-Base, and (5) using the KG embeddings for predicting new causal links.}
 \label{fig:architecture}
\end{figure*}

\section{Method}
The proposed CausalLPBack approach is structured into five primary phases (see Figure \ref{fig:architecture}):
(1) finding and encoding the known causal relations into a causal network, (2) removing the backdoor paths spanning across the nodes in the training and testing set, (3) translating the causal network into a CausalKG, compatible with the causal ontology   
\cite{causalODP}, (4) learning KGE for the CausalKG, and (5) predicting new causal links in the KG. 

\subsection{Causal Network}
A causal network is a graphical model known as a causal Bayesian network, structured as a directed acyclic graph. In this model, nodes symbolize events, and edges represent the causal links between these events. The network, denoted as \(CN = (N^{cn}, E^{cn}, W^{cn})\), comprises nodes \(N^{cn}\), edges \(E^{cn}\), and causal weights \(W^{cn}\) which denote the strength of the causal influence. The direction of each edge in the network indicates the direction of causality, and causal weights assigned to these edges measure the impact of changes in one node on another, using do-calculus.  The network follows the local Markov property, which states that knowing a node's direct causes makes it independent of unrelated effects. 

\subsection{Backdoor path}
A Backdoor (BD) is a non causal path for a given node pair ($N^{cn}_c$, $N^{cn}_e$) in a causal network, where $N^{cn}_c$ is the cause and $N^{cn}_e$ is the effect. It indicates common causes of the node pair. To estimate the true influence of a cause, $N^{cn}_c$, on its effect, $N^{cn}_e$, removing of all the backdoor paths blocks the spurious paths between the cause and effect node pair which may otherwise lead to inflated results.

A set of nodes $BD_i$ is said to satisfy as a backdoor path for a given node pair in a causal event graph if:
\begin{itemize}
    \item no node in $BD_i$ is a child of cause node,$N^{cn}_c$; and 
    \item $BD_i$ blocks every path between cause $N^{cn}_c$ and effect $N^{cn}_e$ containing an incoming edge into the cause. 
\end{itemize}

We considered two sets of BD paths: sufficient BD paths and maximum backdoor paths. The sufficient BD is the minimum set required to satisfy the BD path. The paths with direct incoming edge (parent) into the cause node. The maximum backdoor path is a maximum possible set which includes all the possible BD paths for a given pair.

\subsection{Causal Knowledge Graph}
The process of transforming data from a causal network into a causal knowledge graph (CausalKG) involves several conversions:
\begin{itemize}
    \item Nodes ($N^{cn}$) in the causal network are translated into causal entities (\(N_c\)) in the CausalKG.
    \item Weights (\(W^{cn}\)) in the causal network become causal weights (\(W_c\)) associated with a causal link.
    \item Edges (\(E^{cn}\)) in the causal network are translated into causal links (\(E_c\)), structured as quadruples $<$$n_{cause}$, $r_{causes}$, $n_{effect}$, $w_c$$>$.
\end{itemize}
Beyond these direct conversions, the CausalKG is enriched with additional causal links using relations such as \(causedBy\), \(causesType\), and \(causedByType\), ensuring that the graph adheres to a causal ontology that defines the semantics and structure of these relations. This ontology is rooted in concepts from causal AI, like causal Bayesian networks and do-calculus \cite{causalODP}. For causal link prediction within this framework, the CausalKG employs KG link prediction techniques. Specifically, for causal explanation, the system aims to predict the type of a cause-entity linked to an effect-entity, navigating through a two-hop path that may involve reified relations like \(causedByType\) to simplify direct type predictions. Similarly, for causal prediction, the \(causesType\) relation facilitates direct linkage between a cause-entity and the type of an effect-entity. Additionally, the CausalKG can incorporate extra domain knowledge about the causal entities that might not be explicitly detailed in the original causal network, further enriching the graph’s utility and accuracy.

\subsection{CausalKG Embedding and Link Prediction}
A CausalKG can be transformed into a low-dimensional, continuous latent vector space known as a knowledge graph embedding (KGE), which are useful for downstream tasks like link prediction, triple classification, entity classification, and relation extraction  \cite{wang2017knowledgesurvey}. The CausalLPBack approach employs KGE algorithms to create two types of embeddings for CausalKG: CausalKGE-Base, which lacks causal weights, and CausalKGE-W, which incorporates these weights to produce weighted embeddings \cite{focusE}. These embeddings are created using the Ampligraph\footnote{https://docs.ampligraph.org/en/latest/} library tools.
The process involves training the CausalKGE-Base using causal links without considering the associated weights, while the CausalKGE-W considers these weights, assigning higher probabilities to links with greater weights and treating zero-weighted links as negative samples. Training adjusts the output from the scoring layer based on the causal weights before progressing to the loss layer, thereby modulating the scores to reflect the weighted causal impacts. These embeddings are primarily evaluated for causal link prediction, utilizing KG link prediction techniques. CausalLPBack frames causal link prediction as a KG link prediction task, using the embeddings to identify missing causal links and to facilitate causal explanation and prediction tasks. Causal explanation seeks to determine the causes of effects, whereas causal prediction focuses on the effects of causes. This method, tested using the CLEVRER-Humans causal reasoning benchmark dataset, demonstrates the utility of CausalLPBack in generating a CausalKG and employing its embeddings for effective causal link prediction \cite{mao2022clevrerhumans}.

\begin{figure}
 \centering
 \includegraphics[width=0.45\textwidth,height=0.25\textwidth]{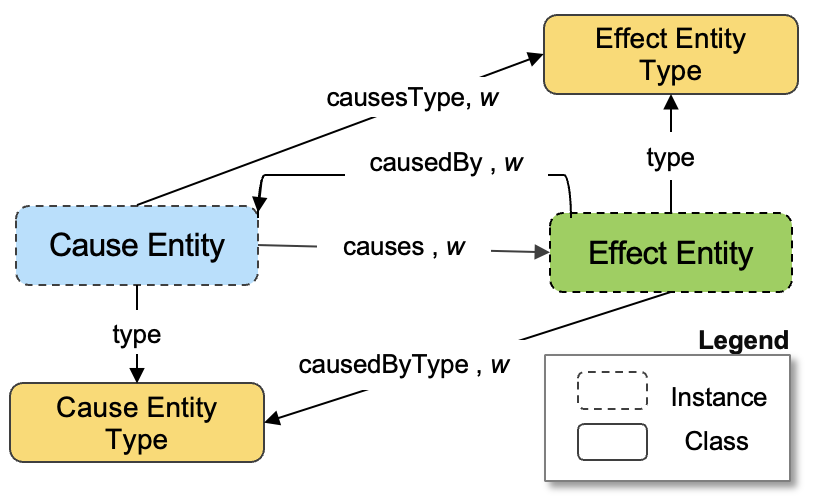}
 \caption{Reified causal relation, \textit{causesType} and \textit{causedByType}. The \textit{causesType} is a reified relation from a cause-entity instance to the type of an effect-entity. The \textit{causedByType} is a reified relation from an effect-entity instance to the type of a cause-entity.}
 \label{fig:reifiedCausalRelation}
\end{figure}

\section{Experiment}
The causal link prediction approach is assessed using KG link prediction, tailored for causal explanation and causal prediction. Specifically, causal explanation involves predicting the type of a cause-entity from a given effect-entity using the causal link format $<$$n_{effect}$, $r_{causedByType}$, $?$, $w$$>$, while causal prediction entails prediction the type of an effect-entity from a given cause-entity using the format 
$<$$n_{cause}, r_{causesType}, ?, w$$>$ (see Figure \ref{fig:reifiedCausalRelation}). These tasks are demonstrated using different CausalKG variations setup and KGE models with the CLEVRER-Humans benchmark dataset for causal reasoning \cite{mao2022clevrerhumans}.

\subsection{Data}
CLEVRER-Humans is a causal reasoning benchmark dataset that enhances the CLEVRER dataset—a simulation of collision events in videos—with human-annotated causal judgments \cite{mao2022clevrerhumans}. The videos feature distinct moving objects, differentiated by shape (e.g., spheres, cubes, cylinders), color (e.g., blue, red, yellow, green, purple, gray, cyan, brown), and material (e.g., metal, rubber). These objects are involved in various events, such as entering, exiting, colliding, moving, hitting, bumping, and rolling. The dataset captures the causal relationships between these events in a Causal Event Graph (CEG), where event descriptions from the videos serve as nodes and directed edges between these nodes indicate causal connections. Human annotators assess these edges on a scale from 1 to 5 to rate the strength of the causal links, where 1 indicates 'not responsible at all' and 5 denotes 'extremely responsible'. This scoring helps determine the perceived impact of one event on another within the video. 

\subsection{Pre-processing the Data}
The first step towards generating a CausalKG for CLEVRER-Humans involves pre-processing the CEGs. 
\subsubsection{Structure of the CEG}
The Causal Event Graphs (CEGs) in the dataset are treated as approximations of a causal network, requiring pre-processing to align with the formal structure of a causal network that consists of directed acyclic graphs representing causal links between nodes. The initial step in pre-processing involves removing edges scored as 1, indicating no causal connection between the nodes involved. After removing the edges with scores 1, the CEGs are checked for cycles. To adhere to the acyclic nature of causal networks, any edges that create cycles within the CEGs are also removed. Additionally, any CEGs lacking sufficient depth (less than two levels from the root to the leaf node) or remaining causal links after these deletions are discarded. This process results in a set of 764 CEGs suitable for further analysis. \subsubsection{Event extraction}
The CLEVRER-Humans dataset categorizes 27 different events, such as collide, enter, exit, and others, into binary and singular events. Binary events involve interactions between two objects and include events like collide, bump, and hit. Singular events involve just one object and include events like enter, exit, and stop. To process these events, information about the event type and the involved objects are extracted from the descriptions in the Causal Event Graph (CEG) using tools like the Berkeley neural semantic parser\footnote{https://pypi.org/project/benepar/} and the NLTK stem lemmatizer\footnote{https://www.nltk.org/index.html}. These help to standardize event labels to their root forms (e.g., 'collide' instead of 'collided'). Nodes describing composite events involving multiple actions are excluded from the analysis to focus only on nodes representing a single event. \subsubsection{Object and object property extraction}
In addition to event types, the analysis of the CLEVRER-Humans dataset involves extracting details about the participating objects, including their color, shape, and material. During this process, some discrepancies in object labeling are identified, such as incorrect color names (e.g., an object labeled as 'gold' instead of 'yellow'). These errors are addressed by normalizing the terms to ensure consistency and accuracy in the dataset's annotations.

\subsection{CausalKG from CLEVRER-Humans} 
A CausalKG is generated from CLEVRER-Humans by encoding the causal information within the CEGs in RDF\footnote{https://www.w3.org/RDF/} format, conformant with the causal ontology. In addition to causal relations, the KG contains information about events along with the participating objects and their characteristics. We use three ontologies to represent information from the CEGs: causal ontology, scene ontology (prefix \textit{so:}), and semantic sensor network ontology (prefix \textit{ssn:}). The causal ontology is used to represent the events (as causal entities), causal relations, and their associated causal weights (i.e. edge score). The scene ontology and sensor ontology are used to represent the additional information about the video, including scenes, objects, and object characteristics \cite{wickramarachchi2021knowledge,taylor2019semantic}. 
Each video is represented as a scene (\textit{so:Scene}) using concepts from the scene ontology. This includes representing and linking the events included in the scene (with the \textit{so:includes} relation), the participating objects (with the \textit{so:hasParticipant} relation), and the object characteristics (with the \textit{ssn:hasProperty} relation) \cite{wickramarachchi2021knowledge}. In total, the CausalKG from CLEVRER-Humans contains \>48K links, 5664 entities, 31 entity types, and 10 relations. 

\subsection{Splitting the Data and Removing the Backdoor Path }
We use a novel dataset splitting approach, Markov-based split, grounded in the local Markov property of causal relations introduced in \cite{CausalDisco}. The Markov-based data split divides each of the causal networks into a train and test set.  The split is based on the local Markov property of a causal network; i.e. for a given direct cause of a node, it is independent of its non-effects.   With the Markov-based data split, the initial train and test sets contains 80\% and 20\% of the total CEGs, respectively. From the 764 CEGs, 612 are in train and 152 are in test set.  The CEGs in the test set are further split into Markov-train and Markov-test set based on the local Markov property of a causal network.   
We remove 1) sufficient backdoor path (minimum), and 2) maximum backdoor path satisfying the backdoor path criteria from the Markov-train and Markov-test sets. The CEG present in the training set, as well as the nodes and relationships within the Markov-train set, are utilized to construct the CausalKG specific to CLEVRER-Humans. This CausalKG is subsequently employed to train the KGE. Similarly, the nodes and relationships within the Markov-test set are utilized to create test links, which are then employed to assess the performance of the KGE. Each dataset split is provided as input to the KGE algorithms, resulting in the generation of both 1) CausalKGE-Base embeddings without causal weights, and 2) CausalKGE-W embeddings with causal weights. These embeddings are intended for use in the task of link prediction, specifically for causal explanation and prediction.

\begin{figure*}
 \centering
\includegraphics[width=\textwidth, height=0.35\textwidth]{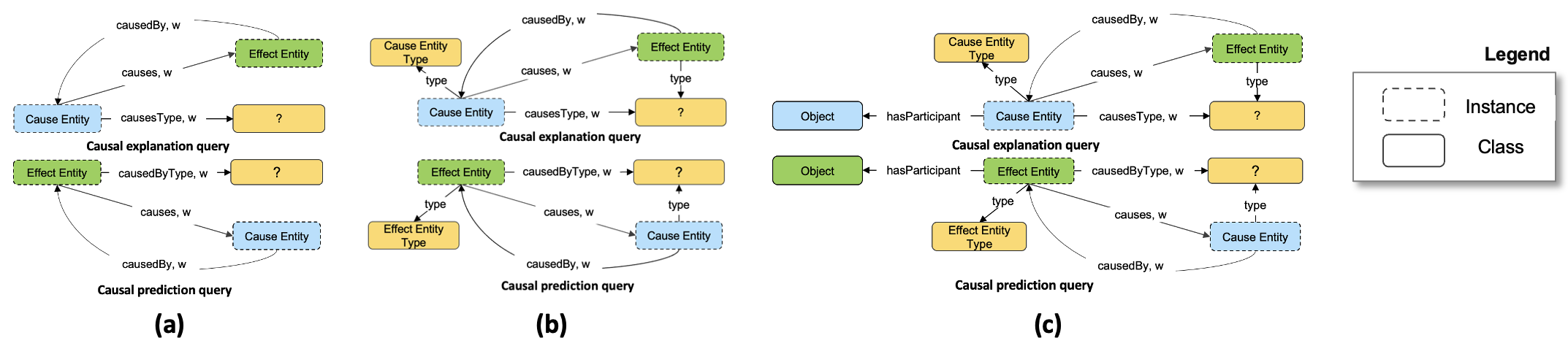}
 \caption{Different CLEVRER-Humans CausalKG structures used for evaluating causal explanation and prediction tasks: (a) subgraph C which consists of links with only causal relations, i.e. \textit{causes}, \textit{causedBy}, \textit{causesType}, and \textit{causedByType}, (b) subgraph CT with causal relations and information about entity types, i.e. \textit{rdf:type}, (c) subgraph CTP with causal relations, entity type relations, and information about the objects that participate in the causal events.} 
 \label{fig:KGstructures}
\end{figure*} 

\subsection{Diversifying the Available Knowledge}

The CausalKGE-Base and CausalKGE-W embeddings are developed and tested across various subgraph structures within the CLEVRER-Humans CausalKG to assess their effectiveness in causal explanation and prediction tasks. These subgraphs vary in complexity and detail, as illustrated in Figure  \ref{fig:KGstructures}. Specifically, three different subgraph structures are defined with increasing levels of expressivity:
(1) graph structure C: includes only basic links with causal relations, (2)
graph structure CT: includes causal relations and types of causal entities, adding a layer of semantic detail, (3) graph structure CTP: the most detailed structure, includes causal relations, entity types, and relationships to objects involved in causal events. For each graph structure, hyper-parameters were carefully optimized to enhance the performance of the corresponding tasks—causal explanation and prediction. The optimized CausalKGE models are then applied to causal link prediction, utilizing established link prediction techniques to evaluate their performance effectively. 

\subsection{Evaluation Metrics}
The evalution of CausalLPBack is performed by following the KG link prediction experiment design for four common KGE algorithms: TransE \cite{TransEbordes2013translating}, DistMult \cite{DistMultyang2014embedding}, HoLE \cite{HolEnickel2016holographic}, and ComplEx \cite{ComplExtrouillon2016complex}. Evaluations performed with weights use FocusE \cite{focusE}. For a given set of causal links $E_c$ in CausalKG, a set of corrupted links $\mathcal{T}'$ are generated by replacing the head $h_c$ or tail $t_c$ of a set of causal links, $<$$h_c, r_c, t_c, w_c$$>$, with another causal entity in the KG. Replacing the head with $h_c' \neq h_c$ results in $<$$h_c', r_c, t_c, w_c$$>$ or replacing the tail with $t_c' \neq t_c$ results in $<$$h_c, r_c, t_c', w_c$$>$. 
The model scores the true link $<$$h_c, r_c, t_c, w_c$$>$ and corrupted links $<$$h_c', r_c, t_c, w_c$$>$, $<$$h_c, r_c, t_c', w_c$$>$ $\in \mathcal{T}'$. The scores are then sorted to obtain the rank of the true link. The filtered evaluation setting and filtered corrupted links $\mathcal{T}'$ are used to exclude the links present in the training and validation set. The overall performance of the models is measured using mean reciprocal rank (MRR) and Hits@K for k = \{1,3,10\}. MRR is the mean over the reciprocal of individual ranks of test links. Hits@k is the ratio of test links present among the top k ranked links. 

\begin{table}[]
\centering
\begin{tabular}{|c|c|c|c|c|}
\hline
\textbf{Algorithm} & \textbf{Subgraph} & \textbf{MRR w/ BD} & \textbf{MRR w/o BDMax} & \textbf{MRR w/o Sufficient} \\ \hline
TransE & C & 0.43 & 0.1864 & 0.1864 \\ \hline
TransE & CT & 0.23 & 0.1859 & 0.1833 \\ \hline
TransE & CTP & 0.30 & 0.2495 & 0.2384 \\ \hline
TransE & Average & {\textbf{0.319}} & {\textbf{0.207}} & {\textbf{0.203}} \\ \hline
DistMult & C & 0.33 & 0.245 & 0.1171 \\ \hline
DistMult & CT & 0.22 & 0.1667 & 0.1658 \\ \hline
DistMult & CTP & 0.18 & 0.2301 & 0.1092 \\ \hline
DistMult & Average & {\textbf{0.243}} & {\textbf{0.214}} & {\textbf{0.131}} \\ \hline
HoLE & C & 0.28 & 0.1649 & 0.0978 \\ \hline
HoLE & CT & 0.18 & 0.1573 & 0.0651 \\ \hline
HoLE & CTP & 0.20 & 0.1279 & 0.0568 \\ \hline
HoLE & Average & {\textbf{0.220}} & {\textbf{0.150}} & {\textbf{0.073}} \\ \hline
ComplEx & C & 0.32 & 0.1094 & 0.0663 \\ \hline
ComplEx & CT & 0.13 & 0.1201 & 0.015 \\ \hline
ComplEx & CTP & {0.12} & 0.2726 & 0.2726 \\ \hline
ComplEx & Average & {\textbf{0.188}} & {\textbf{0.167}} & {\textbf{0.118}} \\ \hline
\end{tabular}
\caption{The MRR scores KGE models with backdoor (w/ BD), maximum backdoor elimination (w/o BDMax), and sufficient backdoor elimination (w/o Sufficient) for causal prediction without weights}
\label{tab:table1}
\end{table}

\begin{table}[]
\centering
\begin{tabular}{|c|c|c|c|c|}
\hline
\textbf{Algorithm} & \textbf{Subgraph} & \textbf{MRR w/ BD} & \textbf{MRR w/o BDMax} & \textbf{MRR w/o Sufficient} \\ \hline
TransE & C & 0.85 & 0.5307 & 0.5307 \\ \hline
TransE & CT & 0.71 & 0.4143 & 0.3796 \\ \hline
TransE & CTP & 0.80 & 0.6006 & 0.5315 \\ \hline
TransE & Average & { \textbf{0.787}} & {\textbf{0.515}} & {\textbf{0.481}} \\ \hline
DistMult & C & 0.37 & 0.3457 & 0.338 \\ \hline
DistMult & CT & 0.44 & 0.3058 & 0.3897 \\ \hline
DistMult & CTP & 0.48 & 0.2877 & 0.272 \\ \hline
DistMult & Average & {\textbf{0.430}} & {\textbf{0.313}} & {\textbf{0.333}} \\ \hline
HoLE & C & 0.36 & 0.3121 & 0.2397 \\ \hline
HoLE & CT & 0.60 & 0.3029 & 0.2605 \\ \hline
HoLE & CTP & 0.51 & 0.3145 & 0.2689 \\ \hline
HoLE & Average & {\textbf{0.490}} & {\textbf{0.310}} & {\textbf{0.256}} \\ \hline
ComplEx & C & 0.37 & 0.3497 & 0.2277 \\ \hline
ComplEx & CT & 0.36 & 0.3287 & 0.1938 \\ \hline
ComplEx & CTP & {0.32} & 0.3245 & 0.3245 \\ \hline
ComplEx & Average & {\textbf{0.350}} & {\textbf{0.334}} & {\textbf{0.249}} \\ \hline
\end{tabular}
\caption{The MRR scores KGE models with backdoor (w/ BD), maximum backdoor elimination (w/o BDMax), and sufficient backdoor elimination (w/o Sufficient) for causal prediction with weights}
\label{tab:table2}
\end{table}

\begin{figure}
\centering
    \includegraphics[width=\textwidth]{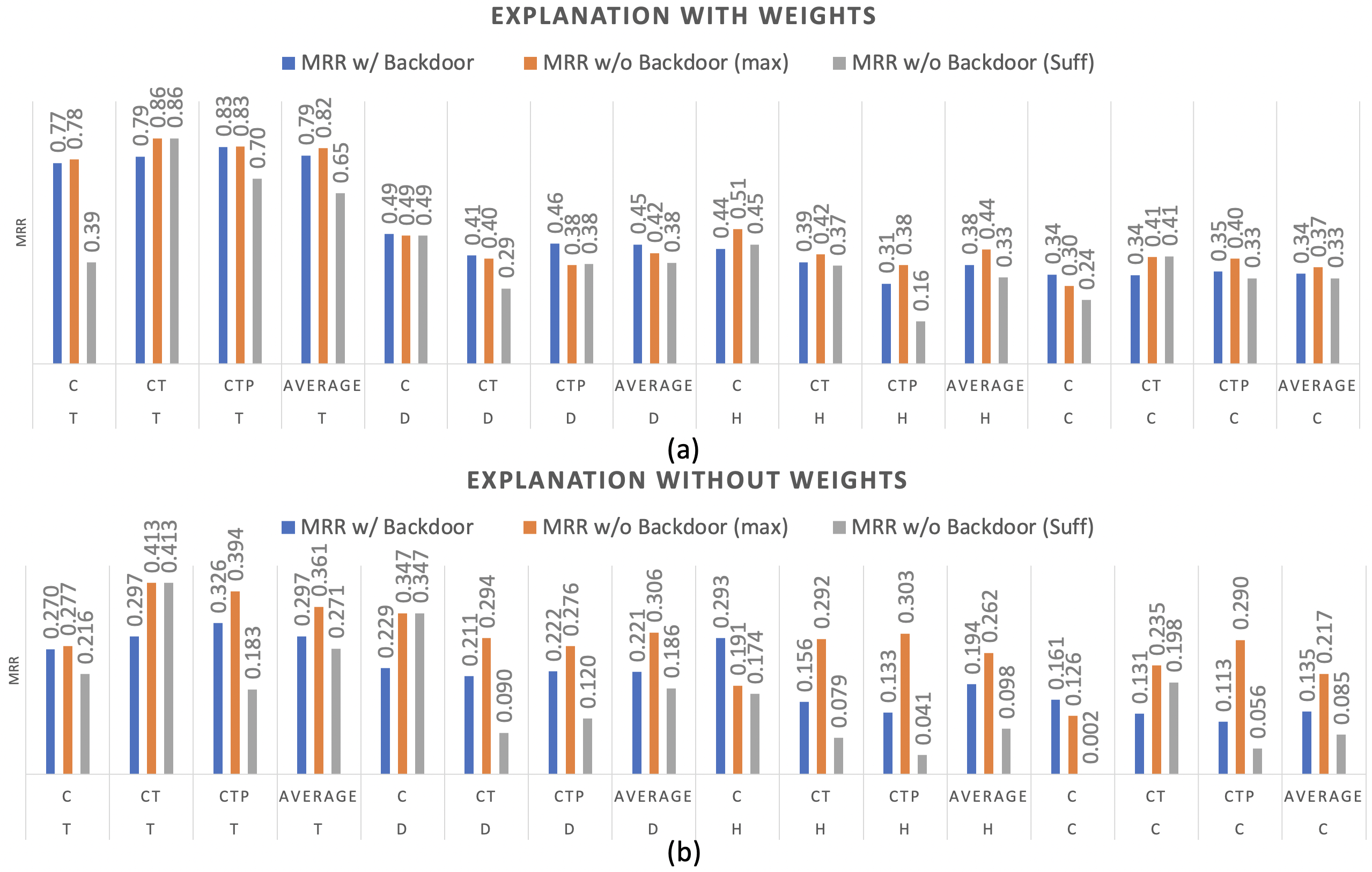}
    \caption{The MRR scores KGE models with backdoor (w/ Backdoor), maximum backdoor elimination (w/o Backdoor (max)), and sufficient backdoor elimination (w/o Backdoor (Suff)) for (a) causal explanation with weights, (b) causal explanation without weights}
    \label{fig:Explanation}
\end{figure}

\begin{figure}
    \centering
    \includegraphics[width=\textwidth, height=0.45\textwidth]{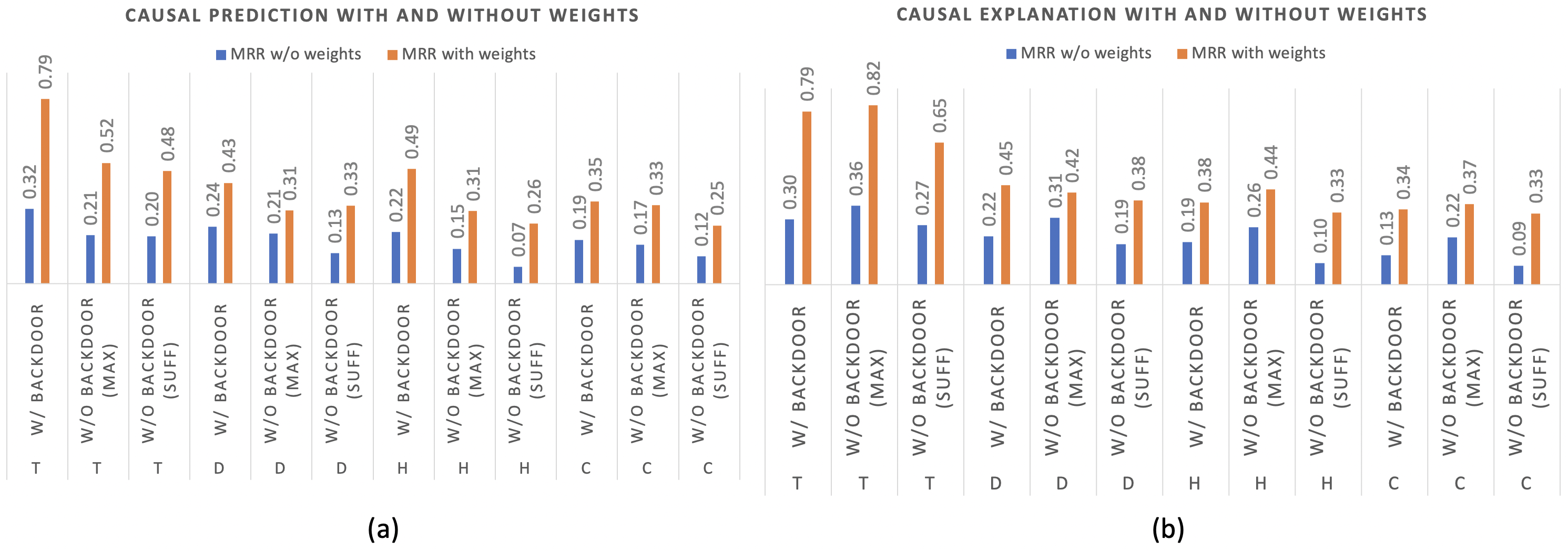}
    \caption{Comparing of MRR for (a) causal prediction, and (b) causal explanation with and without weighted causal relations for with backdoor, without maximum backdoor elimination, and without sufficient backdoor elimination. T refers to TransE, D is DistMult, H is HolE, and C is ComplEx KGE models}
    \label{fig:comparisonWithWithoutWeights}
\end{figure}




\begin{figure}
    \centering
    \includegraphics[width=\textwidth, height=0.5\textwidth]{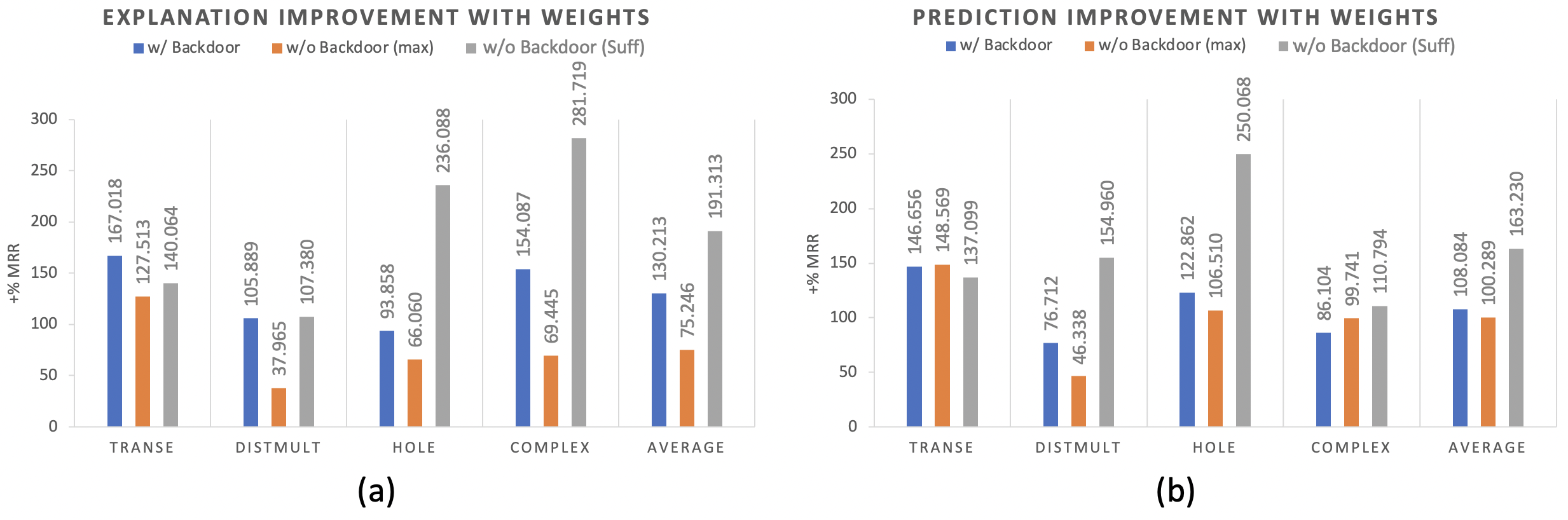}
    \caption{Improvement in +\%MRR for causal links with weights vs without weighted averaged across different CausalKG subgraph structure for (a)causal explanation and (b)causal prediction}
    \label{fig:ImprovementWithWeights}
\end{figure}

\begin{figure}
    \centering
    \includegraphics[width=\textwidth, height=0.5\textwidth]{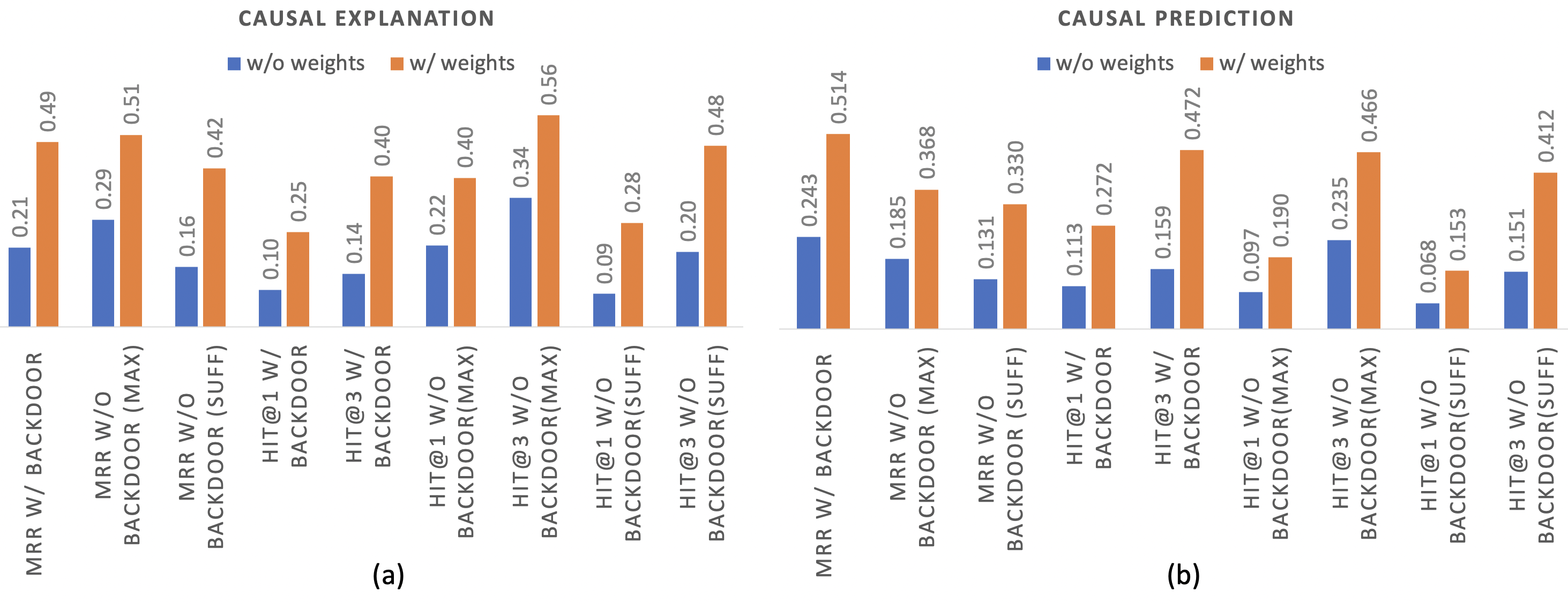}
    \caption{Average performance of MRR and Hits@k across different CausalKG subgraph structure and knowledge graph embedding models with backdoor paths, maximum backdoor elimination (Max), and sufficient backdoor elimination (Suff) for (a)causal explanation, (b)causal prediction}
    \label{fig:AverageComparison}
\end{figure}

\section{Results and Discussion}
CausalLPBack was evaluated on CausalKGs generated (Figure \ref{fig:KGstructures}) from CLEVRER-Humans dataset for the task of causal explanation and prediction. 
CausalLPBack has six trained KGE models:
\begin{enumerate} 
    \item CausalKGE-Base-WithBackdoor 
    \item CausalKGE-W-WithBackdoor
     \item CausalKGE-Base-WithoutMaximumBackdoor 
    \item CausalKGE-W-WithoutMaximumBackdoor 
    \item CausalKGE-Base-WithoutSufficientBackdoor 
    \item CausalKGE-W-WithoutSufficientBackdoor 
\end{enumerate}


The evaluation results demonstrate the influence of the backdoor paths in inflating the performance of causal link prediction as shown in Table \ref{tab:table1}, Table \ref{tab:table2}, Figures \ref{fig:Explanation}, \ref{fig:comparisonWithWithoutWeights}, \ref{fig:ImprovementWithWeights}, and \ref{fig:AverageComparison}. The CausalKGE with the backdoor paths have higher MRRs and Hits@K across the board compared to CausalKGE-SufficientBackdoor and CausalKGE-MaximumBackdoor due to the confounding bias introduced in the link prediction with the presence of the backdoor paths. We observe all across the board the weighted causal relations in the CausalKGE-W outperforms the CausalKGE-Base for all the KGE models, different subgraph structures and varied backdoor elimination. We observe a decrease in performance of MRR score for both weighted and base CausalKGE from WithoutBackdoor, to MaximumBackdoor and SufficientBackdoor for causal prediction.  
The inflation in the MRR and Hits@K metric is higher when backdoor path is compared with sufficient (the minimum paths) backdoor paths vs maximum path elimination. 
In the case of causal explanation, the MRRs and Hit@k demonstrate inflated performance of backdoor when compared with sufficient backdoor elimination.  

\section{Conclusion}
In this paper, we present a novel approach to causal link prediction,\\ CausalLPBack. This approach fills a crucial gap in the current state-of-the-art by integrating the concepts and representation of causality from Causal AI into a knowledge graph (Neuro-Symbolic AI). This method utilizes knowledge graph link prediction to effectively address the confounding bias introduced by backdoor paths in the causal network.  The CausalKGE-W, one of our proposed approaches, outperforms all baseline CausalKGE-Base methods, demonstrating its superiority and innovation. Our results demonstrate that the effective fusion of weighted causal relation and removing the backdoor paths can facilitate the mapping of causal link prediction (i.e., causal AI) to knowledge graph link prediction.  The approach significantly demonstrated the inflated performance of CausalKG due to the bias introduced due to the backdoor paths. In future work, we would like to further study the difference in maximum vs sufficient backdoor elimination with respective to causal explanation vs causal prediction, the role of different subgraph structures,  how does the characteristics of a causal link such as prediction vs explanation play a role in the performance metric. \textit{Additional Supplementary Material\footnote{Additional results, hyper-parameters optimized for the model. The knowledge graph and source code will be shared with the camera-ready for reproducibility https://github.com/utkarshani/CausalLPBack} } 



%
%

%
%
%
\bibliographystyle{splncs04}
\bibliography{sample-base}

\begin{thebibliography}{10}
\providecommand{\url}[1]{\texttt{#1}}
\providecommand{\urlprefix}{URL }
\providecommand{\doi}[1]{https://doi.org/#1}

\bibitem{TransEbordes2013translating}
Bordes, A., Usunier, N., Garcia-Duran, A., Weston, J., Yakhnenko, O.: Translating embeddings for modeling multi-relational data. Advances in neural information processing systems  \textbf{26} (2013)

\bibitem{hassanzadeh2022causalbuilding}
Hassanzadeh, O.: Building a knowledge graph of events and consequences using wikipedia and wikidata. In: Proceedings of the Wiki Workshop at The Web Conference (2022)

\bibitem{heindorf2020causenet}
Heindorf, S., Scholten, Y., Wachsmuth, H., Ngonga~Ngomo, A.C., Potthast, M.: Causenet: Towards a causality graph extracted from the web. In: Proceedings of the 29th ACM international conference on information \& knowledge management. pp. 3023--3030 (2020)

\bibitem{causalODP}
Jaimini, U., Henson, C., Sheth, A.: An ontology design pattern for representing causality. In: 14th Workshop on Ontology Design and Patterns, WOP 2023 (2023)

\bibitem{jaimini2024causalNSAI}
Jaimini, U., Henson, C., Sheth, A.: Causal neuro-symbolic ai: A synergy between causality and neuro-symbolic methods. IEEE Intelligent Systems  \textbf{39}(2) (2024)

\bibitem{CausalDisco}
Jaimini, U., Henson, C., Sheth, A.: Causallp: Learning causal relations with weighted knowledge graph link prediction. arXiv preprint arXiv:2405.02327  (2024)

\bibitem{mao2022clevrerhumans}
Mao, J., Yang, X., Zhang, X., Goodman, N., Wu, J.: Clevrer-humans: Describing physical and causal events the human way. Advances in Neural Information Processing Systems  \textbf{35},  7755--7768 (2022)

\bibitem{HolEnickel2016holographic}
Nickel, M., Rosasco, L., Poggio, T.: Holographic embeddings of knowledge graphs. In: Proceedings of the AAAI conference on artificial intelligence. vol.~30 (2016)

\bibitem{focusE}
Pai, S., Costabello, L.: Learning embeddings from knowledge graphs with numeric edge attributes. arXiv preprint arXiv:2105.08683  (2021)

\bibitem{pearl2009causality}
Pearl, J.: Causality. Cambridge university press (2009)

\bibitem{Whypearl2018book}
Pearl, J., Mackenzie, D.: The book of why: the new science of cause and effect. Basic books (2018)

\bibitem{taylor2019semantic}
Taylor, K.: The semantic sensor network ontology, revamped. In: 18th International Semantic Web Conference (2019)

\bibitem{ComplExtrouillon2016complex}
Trouillon, T., Welbl, J., Riedel, S., Gaussier, {\'E}., Bouchard, G.: Complex embeddings for simple link prediction. In: International conference on machine learning. pp. 2071--2080. PMLR (2016)

\bibitem{wang2017knowledgesurvey}
Wang, Q., Mao, Z., Wang, B., Guo, L.: Knowledge graph embedding: A survey of approaches and applications. IEEE Transactions on Knowledge and Data Engineering  \textbf{29}(12),  2724--2743 (2017)

\bibitem{wickramarachchi2021knowledge}
Wickramarachchi, R., Henson, C., Sheth, A.: Knowledge-infused learning for entity prediction in driving scenes. Frontiers in big Data  \textbf{4},  759110 (2021)

\bibitem{DistMultyang2014embedding}
Yang, B., Yih, W.t., He, X., Gao, J., Deng, L.: Embedding entities and relations for learning and inference in knowledge bases. arXiv preprint arXiv:1412.6575  (2014)

\end{thebibliography}


\end{document}


\section{Figures}

\begin{Figue}
    \includegraphics[width=\textwidth]{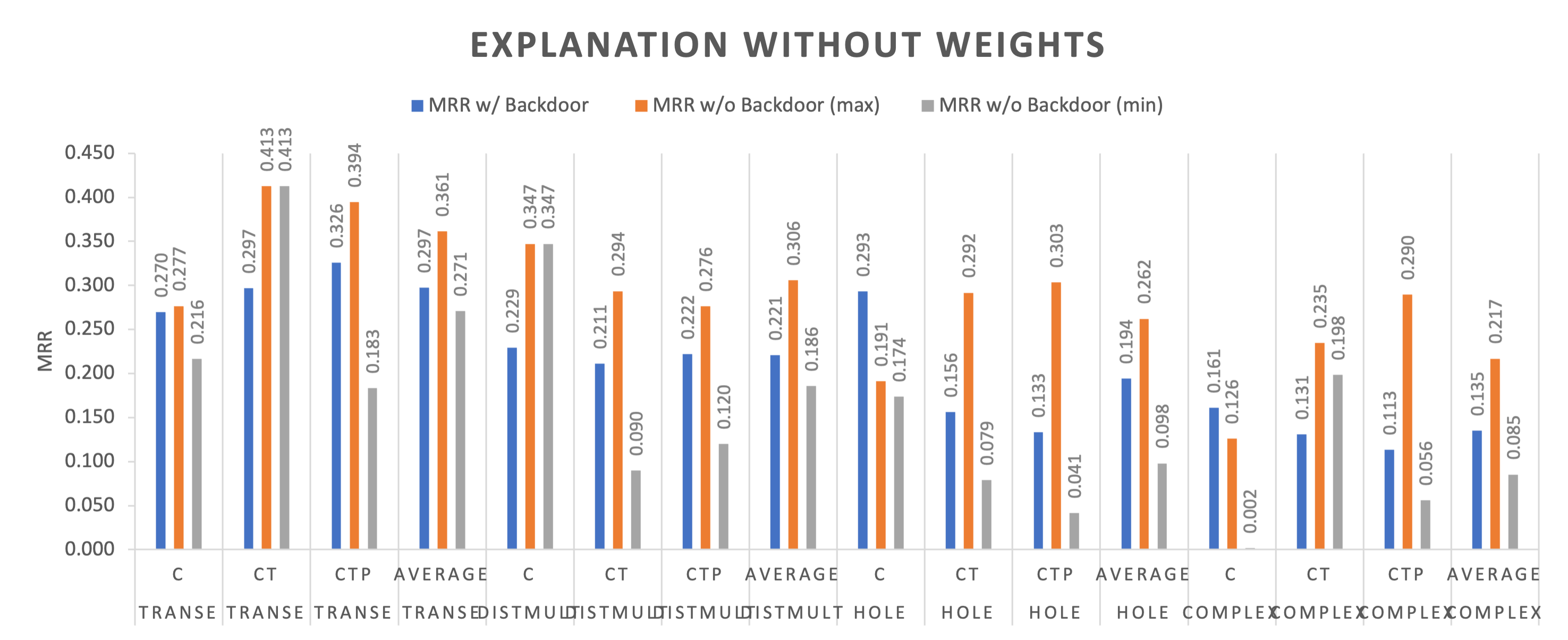}
\end{Figue}
    
\begin{Figue}
    \includegraphics[width=\textwidth]{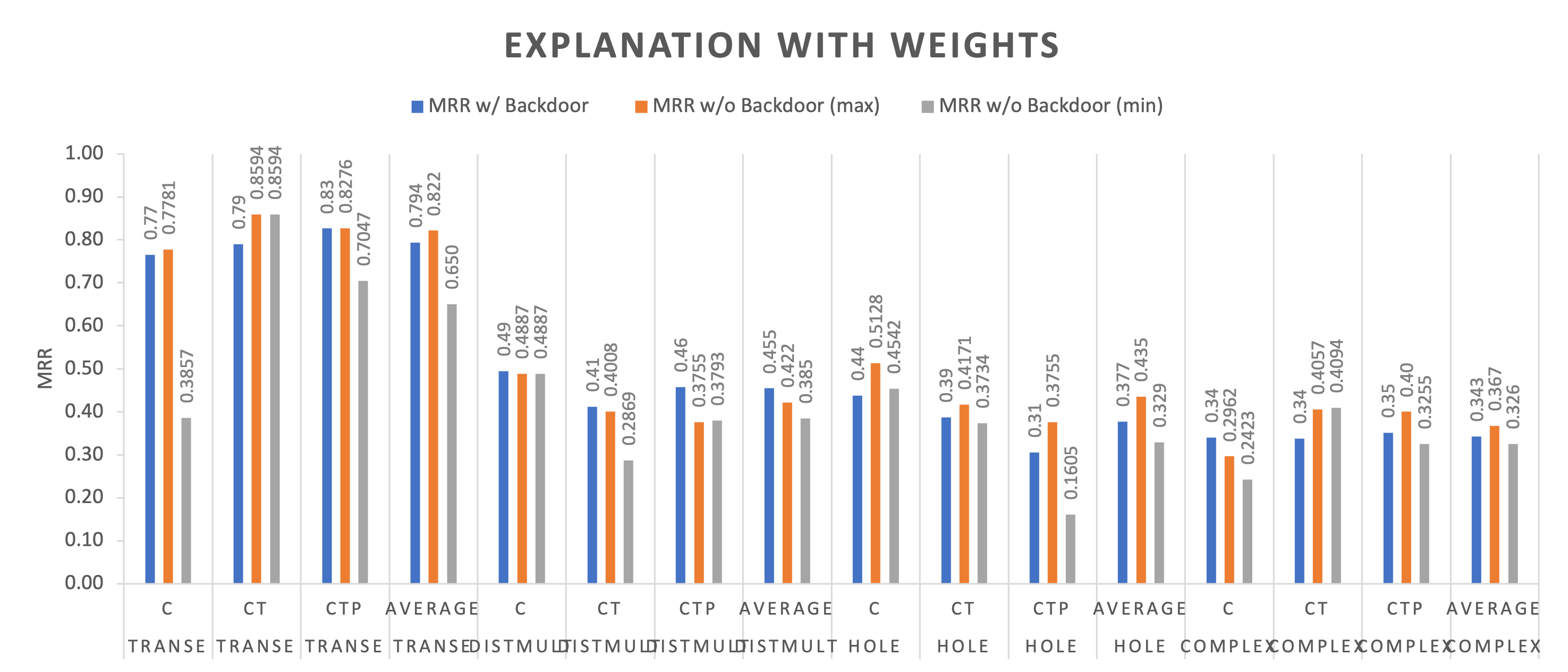}
\end{Figue}

\begin{Figue}
    \includegraphics[width=\textwidth]{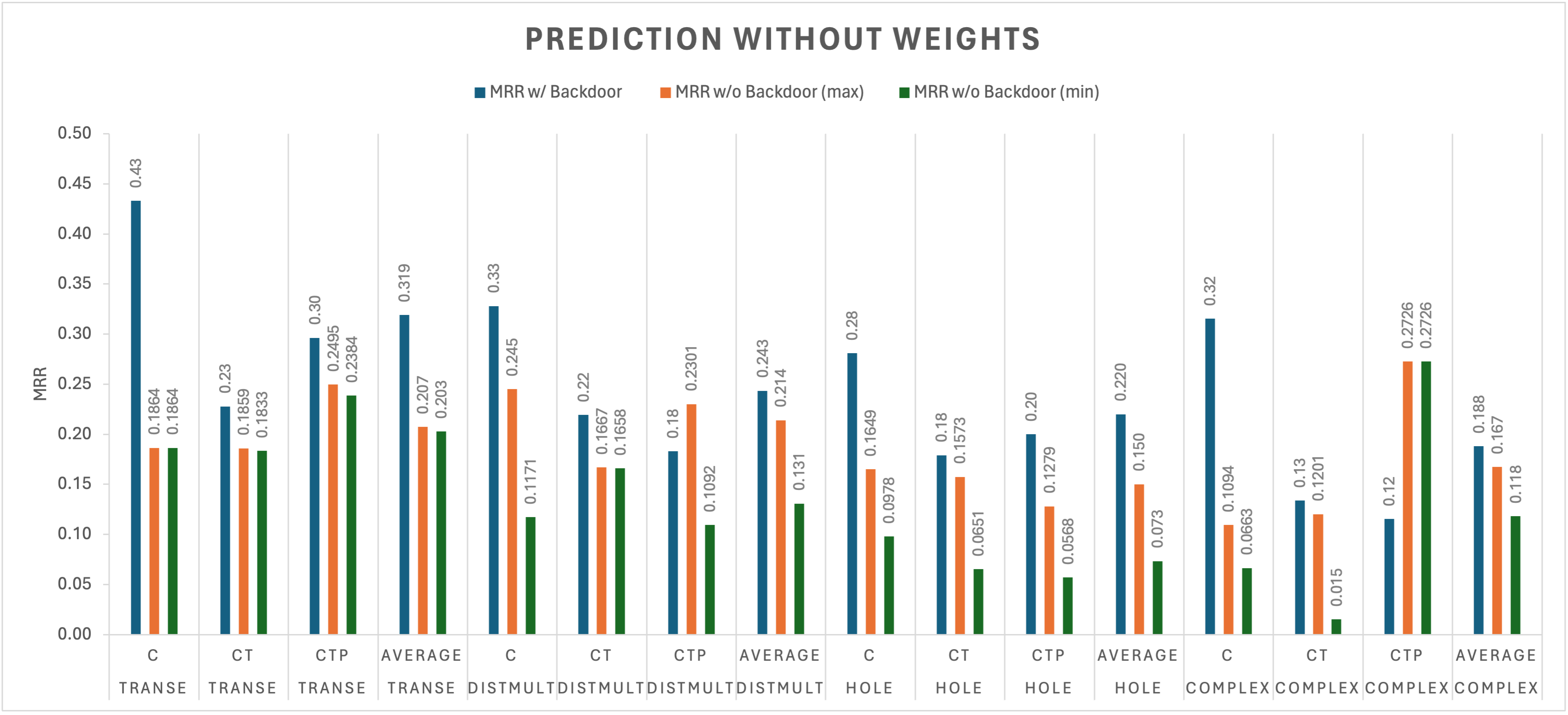}
\end{Figue}

\begin{Figue}
    \includegraphics[width=\textwidth]{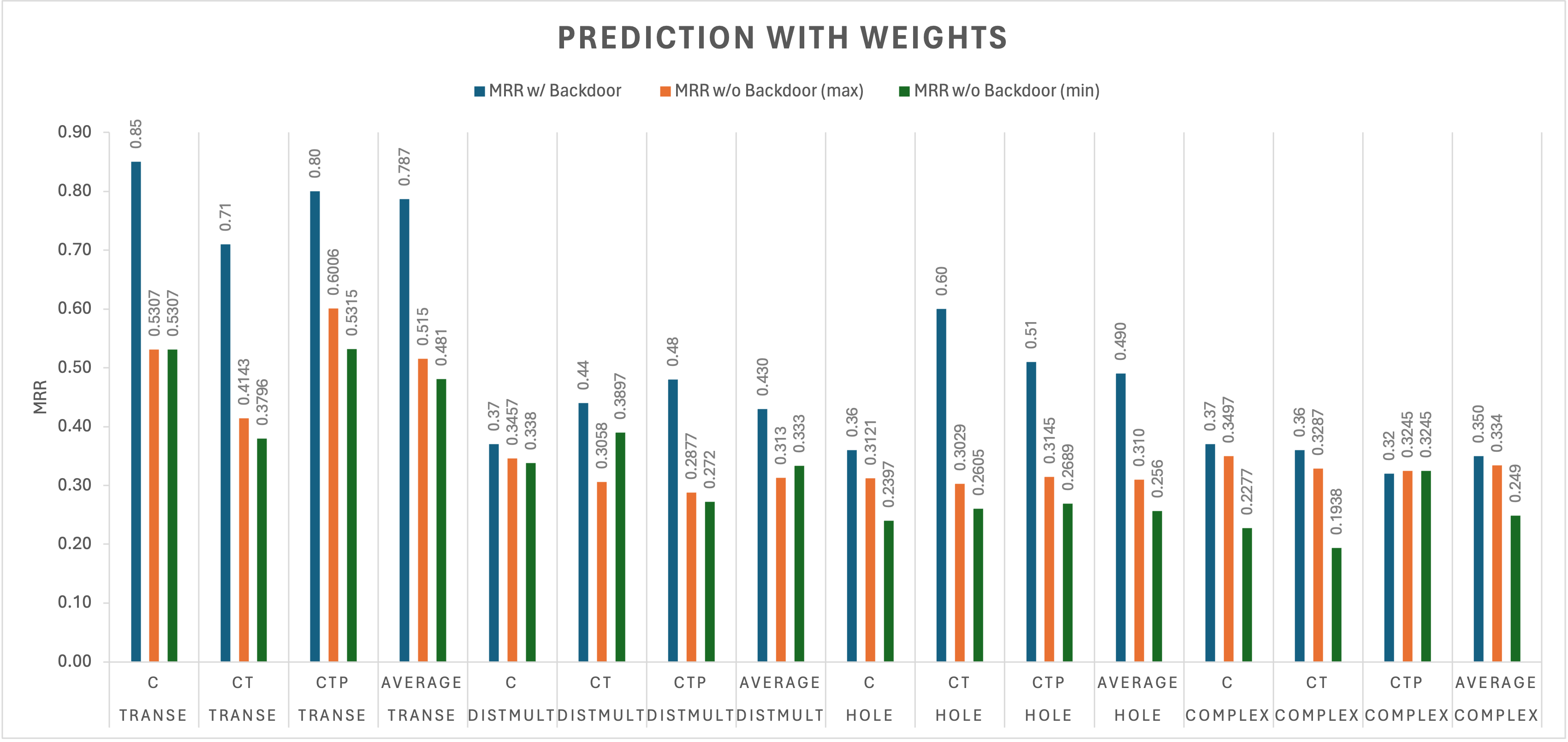}
\end{Figue}

\section{Hyper-parameters}
Hyperparameters for Back2CausalDisco for its six trained KGE models
\begin{landscape}
\begin{table}[htbp]
\centering
\resizebox{\columnwidth}{!}{%
\begin{tabular}{|l|l|l|l|l|l|l|l|l|l|l|}
\hline
KG subgraph structure & Model & Batches Count & Seed & Epochs & K & Eta & Loss & Regularizer & Regularizer Params & Optimizer & Optimizer Params \\ \hline
C & TransE & 100 & 0 & 500 & 200 & 15 & multiclass\_nll & LP & \{"p": 3, "lambda": 1e-05\} & adam & \{"lr": 0.005419994971545206\} \\ \hline
 & DistMult & 100 & 0 & 500 & 200 & 10 & nll & LP & \{"p": 1, "lambda": 1e-05\} & adam & \{"lr": 0.009073539061895694\} \\ \hline
 & HolE & 100 & 0 & 100 & 100 & 5 & multiclass\_nll & LP & \{"p": 3, "lambda": 0.0001\} & adam & \{"lr": 0.006804906722850427\} \\ \hline
 & ComplEx & 100 & 0 & 500 & 100 & 5 & nll & LP &  & adam & \{"lr": 0.006956062741769883\} \\ \hline
CT & TransE & 100 & 0 & 500 & 100 & 15 & multiclass\_nll & LP & \{"p": 1, "lambda": 0.0001\} & adam & \{"lr": 0.0003679829492941821\} \\ \hline
 & DistMult & 100 & 0 & 300 & 200 & 15 & nll &  &  & adam & \{"lr": 0.005470737872602935\} \\ \hline
 & HolE & 100 & 0 & 200 & 200 & 15 & nll &  &  & adam & \{"lr": 0.007873013829029064\} \\ \hline
 & ComplEx & 100 & 0 & 300 & 100 & 10 & nll &  &  & adam & \{"lr": 0.0036637828383358027\} \\ \hline
CTP & TransE & 100 & 0 & 300 & 200 & 15 & nll &  &  & adam & \{"lr": 0.0022465391625126744\} \\ \hline
 & DistMult & 100 & 0 & 300 & 200 & 10 & nll &  &  & adam & \{"lr": 0.009050039489719481\} \\ \hline
 & HolE & 100 & 0 & 500 & 200 & 15 & multiclass\_nll & LP & \{"p": 3, "lambda": 0.0001\} & adam & \{"lr": 0.002808066144674794\} \\ \hline
 & ComplEx & 100 & 0 & 100 & 100 & 10 & nll & LP & \{"p": 3, "lambda": 1e-05\} & adam & \{"lr": 0.0031153621036914864\} \\ \hline
\end{tabular}%
}
\caption{Hyper-parameters for CausalKG-SufficientBackdoor for causal explanation for different subgraph structures and knowledge graph embedding models}
\label{tab:my-table}
\end{table}
\end{landscape}

\begin{landscape}
\begin{table}[htbp]
\centering
\resizebox{\columnwidth}{!}{%
\begin{tabular}{|l|l|l|l|l|l|l|l|l|l|l|}
\hline
\textbf{KG subgraph structure} & \textbf{Model} & \textbf{Batches Count} & \textbf{Seed} & \textbf{Epochs} & \textbf{K} & \textbf{Eta} & \textbf{Loss} & \textbf{Regularizer} & \textbf{Optimizer} \\
\hline
C & TransE & 100 & 0 & 500 & 100 & 15 & multiclass\_nll & LP & adam \\
& DistMult & 100 & 0 & 300 & 200 & 10 & nll & LP & adam \\
& HolE & 100 & 0 & 200 & 200 & 15 & nll &  & adam \\
& ComplEx & 100 & 0 & 500 & 200 & 15 & nll &  & adam \\
\hline
CT & TransE & 100 & 0 & 500 & 100 & 15 & multiclass\_nll & LP & adam \\
& DistMult & 100 & 0 & 500 & 200 & 5 & nll & LP & adam \\
& HolE & 100 & 0 & 500 & 100 & 5 & nll &  & adam \\
& ComplEx & 100 & 0 & 300 & 200 & 15 & nll &  & adam \\
\hline
CTP & TransE & 100 & 0 & 300 & 100 & 5 & multiclass\_nll & LP & adam \\
& DistMult & 100 & 0 & 300 & 100 & 15 & multiclass\_nll &  & adam \\
& HolE & 100 & 0 & 500 & 100 & 5 & nll &  & adam \\
& ComplEx & 100 & 0 & 500 & 200 & 5 & nll & LP & adam \\
\hline
\end{tabular}%
}
\caption{Hyper-parameters for CausalKG-SufficientBackdoor for causal prediction for different subgraph structures and knowledge graph embedding models}
\label{tab:my-table}
\end{table}
\end{landscape}

\begin{landscape}
\begin{table}[htbp]
    \centering
    \resizebox{\linewidth}{!}{%
    \begin{tabular}{cccccccc}
    \toprule
    \textbf{KG subgraph structure} & \textbf{Model} & \textbf{Batches Count} & \textbf{Seed} & \textbf{Epochs} & \textbf{K} & \textbf{Eta} & \textbf{Learning Rate} \\
    \midrule
    \multicolumn{8}{c}{C} \\
    \midrule
    TransE & 100 & 0 & 200 & 200 & 10 & nll & 0.003861276863411193 \\
    DistMult & 100 & 0 & 500 & 200 & 10 & nll & 0.009073539061895694 \\
    HolE & 100 & 0 & 500 & 200 & 15 & multiclass\_nll & 0.002808066144674794 \\
    ComplEx & 100 & 0 & 300 & 200 & 5 & multiclass\_nll & 0.00023104489181310494 \\
    \midrule
    \multicolumn{8}{c}{CT} \\
    \midrule
    TransE & 100 & 0 & 500 & 100 & 15 & multiclass\_nll & 0.0003679829492941821 \\
    DistMult & 100 & 0 & 500 & 100 & 15 & nll & 0.0030317879554692347 \\
    HolE & 100 & 0 & 500 & 200 & 15 & nll & 0.008303703454101804 \\
    ComplEx & 100 & 0 & 500 & 100 & 15 & nll & 0.0030317879554692347 \\
    \midrule
    \multicolumn{8}{c}{CTP} \\
    \midrule
    TransE & 100 & 0 & 500 & 100 & 15 & multiclass\_nll & 0.0003679829492941821 \\
    DistMult & 100 & 0 & 500 & 200 & 15 & nll & 0.00964026132896019 \\
    HolE & 100 & 0 & 300 & 200 & 10 & nll & 0.009050039489719481 \\
    ComplEx & 100 & 0 & 500 & 200 & 5 & nll & 0.008411178371514666 \\
    \bottomrule
    \end{tabular}%
    }
    \caption{Hyper-parameters for CausalKG-BackdoorMaximum for causal explanation for different subgraph structures and knowledge graph embedding models}
    \label{tab:model-parameters}
\end{table}
\end{landscape}

\begin{landscape}
\begin{table}[htbp]
\centering
\begin{tabular}{cccccccc}
\toprule
\textbf{KG subgraph structure} & \textbf{Model} & \textbf{Batches Count} & \textbf{Seed} & \textbf{Epochs} & \textbf{K} & \textbf{Eta} & \textbf{Learning Rate} \\
\midrule
\multicolumn{8}{c}{C} \\
\midrule
TransE & 100 & 0 & 500 & 100 & 15 & multiclass\_nll & 0.0003679829492941821 \\
DistMult & 100 & 0 & 500 & 100 & 5 & nll & 0.008926074507168386 \\
HolE & 100 & 0 & 300 & 200 & 15 & nll & 0.005470737872602935 \\
ComplEx & 100 & 0 & 300 & 200 & 10 & nll & 0.0010649400183823498 \\
\midrule
\multicolumn{8}{c}{CT} \\
\midrule
TransE & 100 & 0 & 200 & 100 & 5 & multiclass\_nll & 0.0065626241383543605 \\
DistMult & 100 & 0 & 500 & 200 & 15 & nll & 0.009568416655679894 \\
HolE & 100 & 0 & 200 & 100 & 10 & nll & 0.009593928946448388 \\
ComplEx & 100 & 0 & 300 & 200 & 10 & nll & 0.0010649400183823498 \\
\midrule
\multicolumn{8}{c}{CTP} \\
\midrule
TransE & 100 & 0 & 500 & 100 & 15 & multiclass\_nll & 0.0003679829492941821 \\
DistMult & 100 & 0 & 300 & 200 & 15 & multiclass\_nll & 0.003988569481398215 \\
HolE & 100 & 0 & 200 & 200 & 15 & nll & 0.00823896555862303 \\
ComplEx & 100 & 0 & 500 & 200 & 5 & nll & 0.008411178371514666 \\
\bottomrule
\end{tabular}
\caption{Hyper-parameters for CausalKG-BackdoorMaximum for causal prediction for different subgraph structures and knowledge graph embedding models}
\label{tab:model-parameters}
\end{table}
\end{landscape}

\begin{landscape}
\begin{table}[htbp]
\centering
\begin{tabular}{cccccccc}
\toprule
\textbf{KG subgraph structure} & \textbf{Model} & \textbf{Batches Count} & \textbf{Seed} & \textbf{Epochs} & \textbf{K} & \textbf{Eta} & \textbf{Learning Rate} \\
\midrule
\multicolumn{8}{c}{C} \\
\midrule
TransE & 100 & 0 & 100 & 200 & 15 & nll & 0.007764913525398746 \\
DistMult & 100 & 0 & 500 & 200 & 5 & nll & 0.008411178371514666 \\
HolE & 100 & 0 & 100 & 200 & 5 & nll & 0.00870793645241166 \\
ComplEx & 100 & 0 & 500 & 100 & 5 & nll & 0.006956062741769883 \\
\midrule
\multicolumn{8}{c}{CT} \\
\midrule
TransE & 100 & 0 & 100 & 200 & 5 & nll & 0.00870793645241166 \\
DistMult & 100 & 0 & 500 & 200 & 15 & nll & 0.00964026132896019 \\
HolE & 100 & 0 & 100 & 200 & 5 & nll & 0.00870793645241166 \\
ComplEx & 100 & 0 & 500 & 100 & 5 & nll & 0.008926074507168386 \\
\midrule
\multicolumn{8}{c}{CTP} \\
\midrule
TransE & 100 & 0 & 200 & 200 & 15 & nll & 0.00823896555862303 \\
DistMult & 100 & 0 & 500 & 200 & 15 & nll & 0.008303703454101804 \\
HolE & 100 & 0 & 500 & 100 & 5 & nll & 0.009595749370048702 \\
ComplEx & 100 & 0 & 300 & 100 & 5 & nll & 0.00259421578363632 \\
\bottomrule
\end{tabular}
\caption{Hyper-parameters for CausalKG-WithBackdoor for causal explanation for different subgraph structures and knowledge graph embedding models}
\label{tab:model-parameters}
\end{table}
\end{landscape}

\begin{landscape}
\begin{table}[htbp]
\centering
\begin{tabular}{cccccccc}
\toprule
\textbf{KG subgraph structure} & \textbf{Model} & \textbf{Batches Count} & \textbf{Seed} & \textbf{Epochs} & \textbf{K} & \textbf{Eta} & \textbf{Learning Rate} \\
\midrule
\multicolumn{8}{c}{C} \\
\midrule
TransE & 100 & 0 & 500 & 100 & 15 & multiclass\_nll & 0.0003679829492941821 \\
DistMult & 100 & 0 & 500 & 200 & 15 & nll & 0.008303703454101804 \\
HolE & 100 & 0 & 500 & 200 & 15 & nll & 0.00964026132896019 \\
ComplEx & 100 & 0 & 500 & 200 & 10 & nll & 0.009073539061895694 \\
\midrule
\multicolumn{8}{c}{CT} \\
\midrule
TransE & 100 & 0 & 300 & 200 & 15 & nll & 0.005470737872602935 \\
DistMult & 100 & 0 & 500 & 100 & 5 & nll & 0.008926074507168386 \\
HolE & 100 & 0 & 500 & 100 & 5 & multiclass\_nll & 0.0035189816388993367 \\
ComplEx & 100 & 0 & 500 & 200 & 5 & nll & 0.008411178371514666 \\
\midrule
\multicolumn{8}{c}{CTP} \\
\midrule
TransE & 100 & 0 & 100 & 200 & 5 & nll & 0.004716476651277433 \\
DistMult & 100 & 0 & 100 & 200 & 5 & nll & 0.0022117885759380717 \\
HolE & 100 & 0 & 200 & 200 & 5 & multiclass\_nll & 0.008342936470924586 \\
ComplEx & 100 & 0 & 300 & 200 & 5 & multiclass\_nll & 0.00023104489181310494 \\
\bottomrule
\end{tabular}
\caption{Hyper-parameters for CausalKG-WithBackdoor for causal prediction for different subgraph structures and knowledge graph embedding models}
\label{tab:model-parameters}
\end{table}
\end{landscape}